\documentclass[letterpaper]{article} 
\usepackage{aaai25}
\usepackage{times}  
\usepackage{helvet}  
\usepackage{courier}  
\usepackage[hyphens]{url}  
\usepackage{graphicx} 
\usepackage{natbib}  
\usepackage{caption} 
\usepackage{algorithm}
\usepackage{algorithmic}
\usepackage{booktabs}
\usepackage{multirow}
\usepackage{amsmath}
\usepackage{amssymb}
\usepackage{mathtools}
\usepackage{amsthm}
\usepackage{diagbox}
\usepackage{bm}
\usepackage[table]{xcolor}
\definecolor{tabhighlight}{HTML}{e5e5e5}

\usepackage{newfloat}
\usepackage{listings}
\DeclareCaptionStyle{ruled}{labelfont=normalfont,labelsep=colon,strut=off} 
\floatstyle{ruled}
\newfloat{listing}{tb}{lst}{}
\floatname{listing}{Listing}
\title{Auto-Regressive Moving Diffusion Models for Time Series Forecasting}

\author{
    Jiaxin Gao\textsuperscript{\rm 1}\textsuperscript{\rm 2}\equalcontrib,
    Qinglong Cao\textsuperscript{\rm 1}\textsuperscript{\rm 2}\equalcontrib,
    Yuntian Chen\textsuperscript{\rm 2}\thanks{Corresponding author}
}
\affiliations{
    \textsuperscript{\rm 1}Shanghai Jiao Tong University, Shanghai, China\\
    \textsuperscript{\rm 2}Ningbo Institute of Digital Twin, Eastern Institute of Technology, Ningbo, Zhejiang, China

    jiaxingao@sjtu.edu.cn; caoql2022@sjtu.edu.cn; ychen@eitech.edu.cn
}

\usepackage{bibentry}

\begin{document}

\maketitle

\begin{abstract}

Time series forecasting (TSF) is essential in various domains, and recent advancements in diffusion-based TSF models have shown considerable promise. However, these models typically adopt traditional diffusion patterns, treating TSF as a noise-based conditional generation task. This approach neglects the inherent continuous sequential nature of time series, leading to a fundamental misalignment between diffusion mechanisms and the TSF objective, thereby severely impairing performance. To bridge this misalignment, and inspired by the classic Auto-Regressive Moving Average (ARMA) theory, which views time series as continuous sequential progressions evolving from previous data points, we propose a novel Auto-Regressive Moving Diffusion (ARMD) model to first achieve the continuous sequential diffusion-based TSF. Unlike previous methods that start from white Gaussian noise, our model employs chain-based diffusion with priors, accurately modeling the evolution of time series and leveraging intermediate state information to improve forecasting accuracy and stability. Specifically, our approach reinterprets the diffusion process by considering future series as the initial state and historical series as the final state, with intermediate series generated using a sliding-based technique during the forward process. This design aligns the diffusion model’s sampling procedure with the forecasting objective, resulting in an unconditional, continuous sequential diffusion TSF model. Extensive experiments conducted on seven widely used datasets demonstrate that our model achieves state-of-the-art performance, significantly outperforming existing diffusion-based TSF models. Our code is available at {https://github.com/daxin007/ARMD}.

\end{abstract}

\section{Introduction}

\begin{figure}[!t]
\centering
\includegraphics[width=1.05\linewidth]{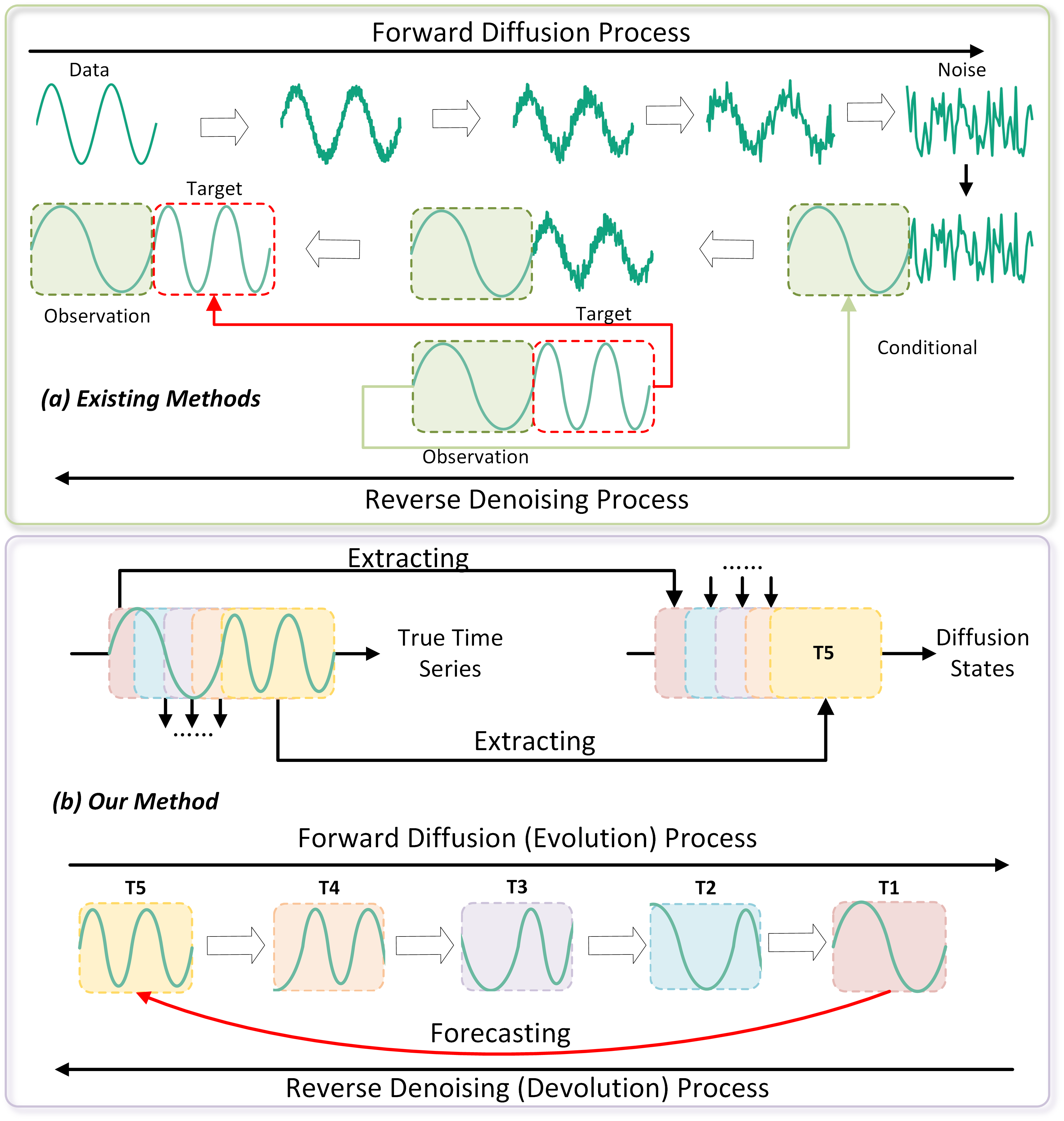} 
\caption{The comparison between (a) existing diffusion-based time series forecasting methods and (b) our methods.}
\vspace{-5mm}
\label{fig:compare}
\end{figure}

Time series forecasting (TSF) plays a pivotal role in various real-world applications, including transportation planning~\cite{fang2022attention}, energy management~\cite{gao2023adaptive, gao2024crossvariablelinearintegratedenhanced}, and financial market analysis~\cite{lopez2023can}. With the development of deep-learning technology, numerous advanced deep-learning TSF models~\cite{piao2024fredformer,wu2023timesnet,haoyietal-informer-2021} have been proposed. Notably, diffusion-based models have emerged as some of the most promising methods in the realm of TSF tasks~\cite{meijer2024rise,yang2024survey}.

Diffusion models have garnered significant attention in computer vision due to their capability to generate high-quality images~\cite{ddpm20,rombach2022high,cao2024teachingvideodiffusionmodel}. These methods typically consist of two main phases: the forward diffusion process, where noise is progressively added to an image until it transforms into white Gaussian noise, and the reverse denoising process, where a denoising network gradually removes noise to reconstruct the original image. While diffusion models are predominantly used in computer vision, recent research has begun exploring their applications in the TSF task.

Current diffusion-based TSF methods~\cite{tashiro2021csdi, alcaraz2022diffusion, yuan2024diffusion} typically follow traditional diffusion patterns, treating TSF as a noise-based conditional generation task. As illustrated in Fig. \ref{fig:compare} (a), the forward diffusion process gradually adds noise to transform the original time series into white Gaussian noise. During forecasting, the historical series serves as a condition, and the reverse denoising process converts the white Gaussian noise into the predicted future time series. While these noise-based conditional diffusion methods have shown some success, they often overlook the continuous sequential nature of time series, leading to a misalignment between the diffusion mechanism and the TSF objective, which significantly suppresses performance. Additionally, by directly diffusing the true series into white Gaussian noise, these methods fail to capture and utilize the valuable intermediate information within the evolution of time series.

To address these limitations, we propose a new continuous sequential diffusion-based model, Auto-Regressive Moving Diffusion (ARMD), for TSF. Drawing inspiration from the principles of the classic TSF theory, Auto-Regressive Moving Average (ARMA), the time series can be viewed as a continuous progression of data points, where each new point is influenced by past data points and includes some random noise. This relationship can be expressed as:
\[\begin{aligned}
x_t &= \phi_1 x_{t-1} + \phi_2 x_{t-2} + \dots + \phi_p x_{t-p} \\
&\quad + \theta_1 \epsilon_{t-1} + \theta_2 \epsilon_{t-2} + \dots + \theta_q \epsilon_{t-q} + \epsilon_t,
\end{aligned}\]
where $x_t$ represents the value at time step $t$, and $\epsilon_t$ represents the random noise. Our model fully leverages the inherent features of time series, which exhibit continuous sequential evolution and form a chain of dependencies characterized by prior information. Unlike previous diffusion methods that focus solely on denoising, ARMD is designed to simulate and learn the underlying evolution of the series.

As illustrated in Fig. \ref{fig:compare} (b), our proposed ARMD introduces a novel approach where the future series is progressively diffused into the historical series during the forward diffusion/evolution process. In contrast, the reverse denoising/devolution process leverages the historical series to forecast the future series. This approach conceptualizes the evolution of a time series as a diffusion process, where each intermediate series at a given time step represents a specific state within this process. Consequently, the forward diffusion process can also be described as a forward evolution process of the time series, and the traditional reverse denoising process is reinterpreted as a reverse devolution process. Unlike conventional methods that introduce noise to generate intermediate states from the original series, ARMD derives intermediate states by sliding the series according to the diffusion steps. This results in a deterministic process that eliminates the uncertainty typically associated with noise. These deterministic intermediate series are then employed for model training. In the reverse devolution process, a linear-based devolution (denoising) network is utilized to devolve the series through a distance-based method. Specifically, the devolution network, equipped with a linear module, estimates the distance between an intermediate state (series) and the target series. It then adaptively adjusts the weighting of the intermediate state and the estimated distance based on the time step, ultimately generating the prediction for the target series and its evolution trend.

During the final sampling/forecasting phase, the model begins with the historical time series and gradually transforms it into the future time series. This approach aligns the sampling process with the ultimate prediction objective, removing the need for conditional generation. Consequently, our method improves stability during both training and sampling, resulting in enhanced forecasting performance.

\begin{itemize}

\item Inspired by classical time series theory, specifically ARMA, and viewing the evolution of time series as an entire diffusion process, we propose the Auto-Regressive Moving Diffusion (ARMD) model as the first continuous sequential diffusion-based TSF model.

\item By employing a simple sliding operation, our method leverages the valuable intermediate information of the time series as intermediate states within the diffusion process. This approach successfully aligns the diffusion mechanism with the true evolution of time series, naturally leading to improved performance.

\item  The proposed ARMD is extensively evaluated on the common TSF benchmarks, and the experimental results demonstrate that our method achieves state-of-the-art forecasting performance.

\end{itemize}



\section{Related work}
Traditional TSF models, such as ARMA and ARIMA \cite{box2015time}, are based on statistical methods that assume linear relationships between past and present observations to identify patterns. Recently, deep-learning models have gained prominence in TSF applications. Temporal convolution networks (TCNs) are a significant branch of deep-learning models, with several TCN-based models \cite{wu2023timesnet,Sen2019ThinkGA, gao2023clearforecastcontrastivelearninghighpurity} leveraging convolution layers to learn temporal dependencies. Transformer-based models \cite{wen2022transformers,xu2024survey} are also widely used in TSF. Among them, PatchTST \cite{nie2022time} segments the time series into multiple tokens and uses an attention module to learn relationships between them. Client \cite{gao2023client} and iTransformer \cite{liu2023itransformer} apply the attention mechanism on the inverted dimensions to capture multivariate correlations. Additionally, DLinear \cite{Zeng2022AreTE} demonstrates the efficacy of linear models in TSF tasks. 


In the realm of diffusion-based TSF models, TimeGrad \cite{timegrad} is a notable denoising diffusion model that generates future values in an auto-regressive manner, effective for short-term forecasting but prone to error accumulation and slow inference for long-term predictions. CSDI \cite{tashiro2021csdi} avoids auto-regressive inference and incorporates a self-supervised strategy through additional input masking. SSSD \cite{alcaraz2022diffusion} builds on CSDI by replacing Transformers with a structured state space model, addressing quadratic complexity issues. However, both CSDI and SSSD face the boundary disharmony problem \cite{lugmayr2022repaint}. TimeDiff \cite{shen2023non} addresses these limitations with additional inductive biases specifically designed for time series data. D3VAE \cite{li2022generative} leverages a coupled diffusion probabilistic model to augment data, integrates multi-scale denoising score matching for improved accuracy, and enhances stability through the disentanglement of multivariate latent variables. TSDiff \cite{kollovieh2024predict} employs a self-guidance mechanism for conditioning during inference without altering the training procedure, excelling in forecasting, refinement, and synthetic data generation tasks. mr-Diff \cite{shen2024multi} employs seasonal-trend decomposition to extract trends during forward diffusion and follows a non-autoregressive denoising process. Similarly, MG-TSD \cite{fan2024mg} leverages intrinsic granularity levels within the data as intermediate diffusion targets, achieving superior forecasting performance. Diffusion-TS \cite{yuan2024diffusion} is an innovative framework that generates high-quality multivariate time series samples using an encoder-decoder Transformer with disentangled temporal representations, effectively extracting essential sequential information from noisy inputs.

\section{Preliminary}
\textbf{Time Series Forecasting.} Given a historical time series ${X_{-L+1:0}}$, where $L$ denotes the series length. The objective of TSF is to forecast the future values of the same series, denoted as ${X_{1:T}}$, where $T$ represents the number of time steps to forecast. ${X_{-L+1:T}}$ can be either a univariate time series or a multivariate time series.

\noindent\textbf{Diffusion Models.}
The traditional diffusion models first progressively add noise \cite{ddpm20} to the original data ${X^{0}}$ through a forward diffusion process, producing noised data ${X^{T}}$. Then the diffusion process is reversed to reconstruct the original data or generate the new data. The intermediate state in each step of diffusion can be calculated based on the original data ${X^{0}}$ and the random noise applied at each step. Trained to predict the added noise or the intermediate state at each step, diffusion models generate new instances by iterative sampling from white Gaussian noise. Diffusion models have been widely applied in diverse fields and have demonstrated notable advantages in generation quality \cite{croitoru2023diffusion}. 


\noindent\textbf{Diffusion Models for TSF in ARMD.} In our proposed ARMD, TSF is aligned with diffusion models by associating the time series with states in the diffusion process, and their relationship is further discussed in the supplemental materials. The future series ${X_{1:T}}$ would serve as the initial state of the diffusion process. Following the notation convention in \cite{shen2023non}, the initial state is further indicated as ${X^{0}_{1:T}}$. Conversely, the historical time series ${X_{-L+1:0}}$ is the final state of the diffusion process, which is further denoted as ${X^{T}_{-L+1:0}}$. Here, the upper index of $X$ indicates its state in the diffusion process, while the lower index represents the time steps it encompasses. In ARMD, the length of the historical series matches that of the future series, allowing the historical series to be defined as ${X^{T}_{-T+1:0}}$. The intermediate state corresponds to a series transitioning from the future series to the historical series, denoted as ${X^{t}_{1-t:T-t}}$.

\begin{figure*}[!t]
\centering
\vspace{-2mm}
\includegraphics[width=0.8\linewidth]{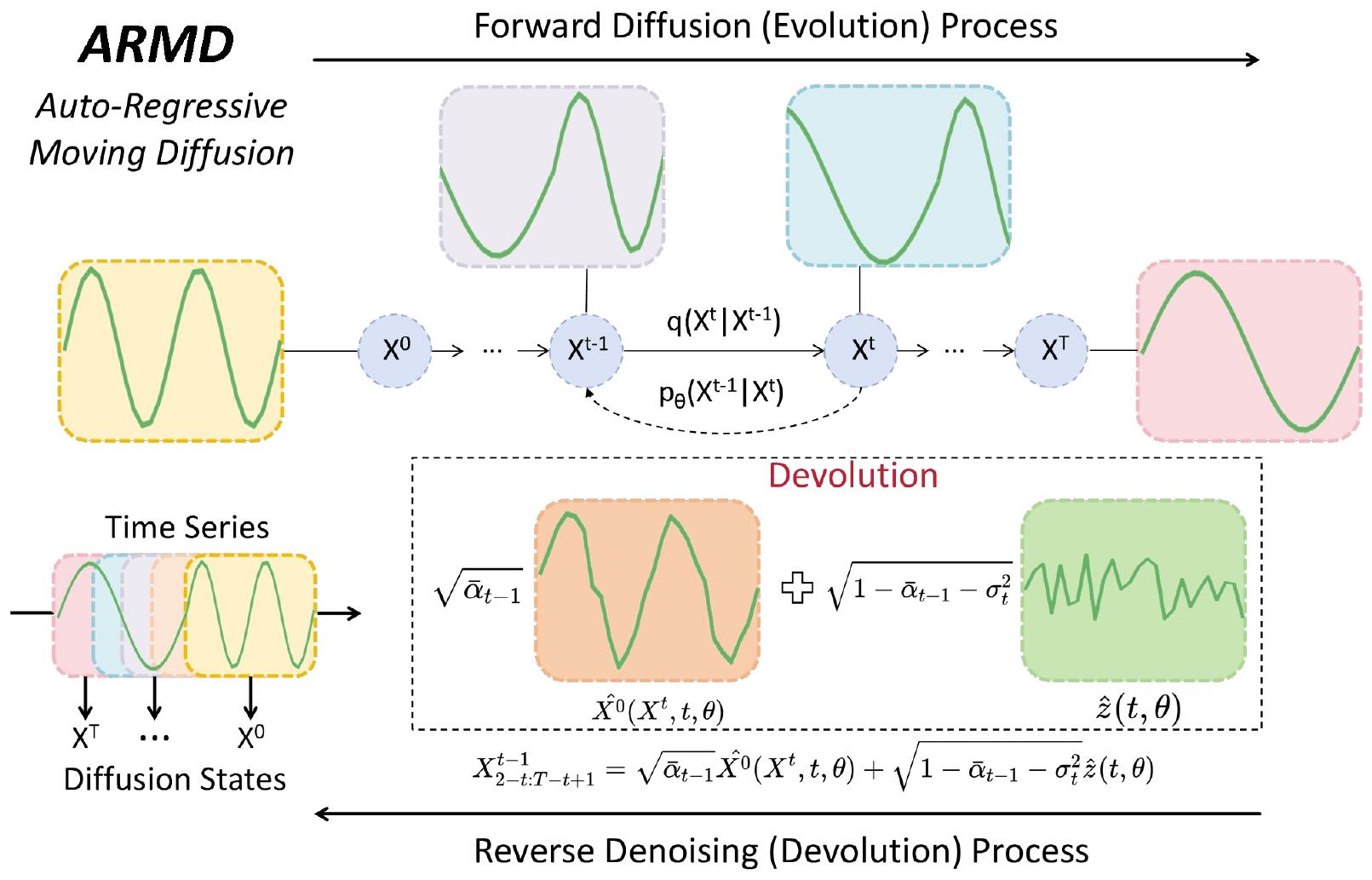} 
\caption{Illustration of the diffusion process in ARMD. During the forward diffusion process, the future series is progressively diffused into the historical series. Conversely, the reverse denoising process utilizes the historical series to iteratively generate/forecast the future series. For clarity, the series $X$ is only annotated with an upper index to indicate its state.}
\vspace{-4mm}
\label{fig:ARMD}
\end{figure*}

\section{Proposed Model}
In this section, we propose Auto-Regressive Moving Diffusion (ARMD), a novel continuous sequential diffusion-based model for TSF. An overview of our proposed model is depicted in Fig. \ref{fig:ARMD}.

\subsection{Forward Diffusion (Evolution) of ARMD}
In ARMD, the evolution of a time series is conceptualized as a diffusion process. Here, the future series ${X^{0}_{1:T}}$ serves as the initial state of the forward diffusion (evolution) process. In contrast, the historical series ${X^{T}_{-T+1:0}}$ represents the final state, as opposed to white Gaussian noise commonly used in traditional diffusion models. Unlike conventional approaches that progressively add noise, the intermediate state ${X^{t}_{1-t:T-t}}$ in ARMD corresponds to sliding the ${X^{0}_{1:T}}$ by $t$ steps, approaching the historical series ${X^{T}_{-T+1:0}}$. This approach leverages the properties of time series, where each intermediate state in the diffusion process reflects an intermediate series in the time series evolution. The process where ${X^{t-1}_{2-t:T-t+1}}$ is diffused to ${X^{t}_{1-t:T-t}}$ represents a single-step movement towards the historical series, as illustrated in Fig. \ref{fig:ARMD}, akin to the $q$ process in denoising diffusion probabilistic model (DDPM) \cite{ddpm20}. The process can be formally expressed as:

\begin{equation}
\begin{aligned}
X^{t}_{1-t:T-t} = \mathrm{Slide}({X^{t-1}_{2-t:T-t+1}}, 1),
\end{aligned}
\end{equation}
where $\mathrm{Slide}(X, k)$ denotes the $k$-step movement of the series window $X$ towards the historical series. Furthermore, at any time step $t$, $X^{t}_{1-t:T-t}$ can be directly obtained using ${X^{0}_{1:T}}$, similar to the process in DDPM. According to the fundamental equation of DDPM in Equation (\ref{eq:x0_samples_xt}), the $t$-step forward process can be rewritten as: 

\begin{align}
X^{t}_{1-t:T-t}=\mathrm{Slide}({X^{0}_{1:T}}, t)=\sqrt{\bar{\alpha}_t}X^{0}_{1:T}+\sqrt{1-\bar{\alpha}_t}z^{t},
\label{eq:x0_zt_xt}
\end{align}
where $z^{t}$ represents the evolution trend from the series $X^{0}_{1:T}$ to $X^{t}_{1-t:T-t}$, functionally similar to the noise added in the original DDPM. Given that each time step of $X^{t}_{1-t:T-t}$ is deterministic, $z^{t}$ can be calculated as:
\begin{equation}
z^t=({\sqrt\frac{1}{\bar{\alpha}_t}}X^{t}_{1-t:T-t}-X^{0}_{1:T}) / \sqrt{\frac{1}{\bar{\alpha}_t}-1}. 
\label{eq:zt1}
\end{equation}
Here, $z^{t}$ serves as the ground truth for the optimization objective at each time step. In this diffusion scheme, the maximum number of diffusion steps $T$ equals the length of the series to be predicted.

\subsection{Reverse Denoising (Devolution) of ARMD}
The reverse process in ARMD utilizes the historical series ${X^{T}_{-T+1:0}}$ to iteratively generate (forecast) the future series ${X^{0}_{1:T}}$. At each devolution step, the linear-based devolution network $R(.)$ predicts the evolution trend $z^{t}$, which is used to devolve ${X^{t}_{1-t:T-t}}$ to ${X^{t-1}_{2-t:T-t+1}}$. Given an intermediate state (series) ${X^{t}_{1-t:T-t}}$ and the diffusion step $t$, $R(.)$ predicts $\hat{X^{0}}(X^{t},t,\theta)$ for the future series ${X^{0}_{1:T}}$. More specifically, within the devolution network $R(.)$, a linear module first provides a prediction of the distance $D$ from the input ${X^{t}_{1-t:T-t}}$ to ${X^{0}_{1:T}}$:
\begin{equation}
\begin{aligned}
D = \mathrm{Linear}({X^{t}_{1-t:T-t}}).
\end{aligned}
\end{equation}

Then, the diffusion step $t$ is used to adaptively balance the interaction between the distance prediction $D$ and the input ${X^{t}_{1-t:T-t}}$, enabling the network to produce more accurate predictions. Particularly, as $t$ decreases, the input ${X^{t}_{1-t:T-t}}$ becomes closer to the target ${X^{0}_{1:T}}$, so the model's output should increasingly resemble the input, placing less emphasis on $D$. Conversely, as $t$ increases, the model places greater reliance on $D$. This adaptive balancing can be mathematically expressed as:
\begin{equation}
\begin{aligned}
\hat{X^{0}}(X^{t},t,\theta) = \frac{W(t)*{X^{t}_{1-t:T-t}} + (1-bW(t))*D}{(1+cW(t))^{d}},
\label{equ:reverse_0}
\end{aligned}
\end{equation}
where $W(t)$ represents a weight coefficient that decreases as $t$ increases, ranging from 0 to 1. We initialize $W(t)$ with the predefined coefficients ${\bar{\alpha}_t}$ of DDPM, and it is updated along with the linear module's parameters during training. The hyper-parameters $b$, $c$, and $d$ are leveraged to balance the interaction between the distance prediction $D$ and the input ${X^{t}_{1-t:T-t}}$. To increase sample diversity and prevent over-fitting, a small deviation is added to the input of $R(.)$ during the training process. More details concerning these hyper-parameters are shown in the supplemental materials.  

After obtaining the prediction $\hat{X^{0}}(X^{t},t,\theta)$, the predicted evolution trend $\hat{z}(t, \theta)$ can be calculated as:

\begin{equation}
\hat{z}(t, \theta)=({\sqrt\frac{1}{\bar{\alpha}_t}}X^{t}_{1-t:T-t}-\hat{X^{0}}(X^{t},t,\theta)) / \sqrt{\frac{1}{\bar{\alpha}_t}-1}. 
\label{eq:zt}
\end{equation}

With the ground truth $z^{t}$ calculated in Equation (\ref{eq:zt1}) and the prediction $\hat{z}(t, \theta)$, the training objective is formulated as:

\begin{equation}
\begin{aligned}
L_{\theta} = \mathrm{E}_{t}[|z^{t}-\hat{z}(t, \theta)|].
\label{equ:loss}
\end{aligned}
\end{equation}


\subsection{Sampling/Forecasting of ARMD}
In the sampling phase, starting from the historical series ${X^{T}_{-T+1:0}}$, ARMD iteratively generates the future series ${X^{0}_{1:T}}$, which successfully aligns the sampling process with the TSF objective, making the model an unconditional diffusion TSF model. The method follows the sampling approach from DDIM \cite{song2020denoising}, replacing the predicted noise $\epsilon_\theta(x_t, t)$ with the predicted evolution trend $\hat{z}(t, \theta)$. The sampling process from $t$ to $t-1$ (akin to the $p$ process in DDPM/DDIM) can be expressed as:
\begin{equation}
\begin{aligned}
X^{t-1}_{2-t:T-t+1} = & \sqrt{\bar{\alpha}_{t-1}} \left( \frac{X^{t}_{1-t:T-t} - \sqrt{1 - \bar{\alpha}_t} \hat{z}(t, \theta)}{\sqrt{\bar{\alpha}_t}} \right) \\ 
& + \sqrt{1 - \bar{\alpha}_{t-1}- \sigma_t^2} \hat{z}(t, \theta) + \sigma_t \epsilon_t,
\label{equ:reverse_origin}
\end{aligned}
\end{equation}
where $\hat{z}(t, \theta)$ is the predicted evolution trend at the time step $t$, and ${\epsilon_t} \sim \mathcal{N}(\mathbf{0}, \mathbf{I})$. 
Since the series evolution in ARMD is deterministic, we remove the noise term $\sigma_t \epsilon_t$. The content within the parentheses in the first term is actually $\hat{X^{0}}(X^{t},t,\theta)$. Thus, the simplified equation is:
\begin{equation}
\begin{aligned}
X^{t-1}_{2-t:T-t+1} &=\sqrt{\bar{\alpha}_{t-1}} \hat{X^0}(X^{t},t,\theta) \\
&+\sqrt{1 - \bar{\alpha}_{t-1} -\sigma_t^2} \hat{z}(t, \theta).
\end{aligned}
\end{equation}

To accelerate sampling, $k$ steps can be skipped at each iteration, and the process can be further inferred as:
\begin{equation}
\begin{aligned}
X^{t-k}_{1-t+k:T-t+k} & = \sqrt{\bar{\alpha}_{t-k}} \hat{X^0}(X^{t},t,\theta) \\ 
&+ \sqrt{1 - \bar{\alpha}_{t-k} -\sigma_t^2} \hat{z}(t, \theta).
\label{equ:reverse_k}
\end{aligned}
\end{equation}

For better understanding, the training and sampling procedures are detailed in Algorithm \ref{alg:training} and Algorithm \ref{alg:sampling}.

\begin{algorithm}[!t]
\caption{Training.}\label{alg:training}
\begin{algorithmic}[1] 
\REQUIRE
Maximum number of diffusion steps $T$, which also represents the length of the historical/future series; Predefined coefficients $\bar{\alpha}_{0:T}$.
\REPEAT
    \STATE Sample $X_{1:T}^0$ from the training set;
    \STATE Sample $t\sim \mathrm{Uniform}(\{1,2,\dots,T\})$;
    \STATE Generate the diffused sample $X^{t}_{1-t:T-t}$ using Equation (\ref{eq:x0_zt_xt}), and calculate the evolution trend $z^t$ using Equation (\ref{eq:zt1});
    \STATE Use the devolution network $R(.)$ to generate the predicted sample $\hat{X^{0}}(X^{t},t,\theta)$ using Equation (\ref{equ:reverse_0}), and obtain the predicted evolution trend $\hat{z}(t, \theta)$ using Equation (\ref{eq:zt});   
    \STATE Calculate the loss $L_{\theta}$ using Equation (\ref{equ:loss});
    \STATE Update the devolution network $R(.)$ of ARMD by taking a gradient descent step on $\nabla_{\theta}L$;
\UNTIL {converged}.
\end{algorithmic}
\end{algorithm}

\section{Experiments}

\begin{table*}[!t]
    \centering
        \caption{Result comparisons of multivariate series forecasting with diffusion-based TSF models. The best results are highlighted in bold. The ``Best Count" column indicates the times of achieving the best result.}
        \vspace{-2mm}
    \label{tab:main-multi}
    \setlength{\tabcolsep}{2.0pt}
    \renewcommand{\arraystretch}{1.1}
    	\scalebox{1.0}{
    \begin{tabular}{lcccccccccc}
        \toprule
       \multicolumn{1}{c}{Methods} & Metric & Solar Energy & ETTh1 & ETTh2 & ETTm1 & ETTm2 & Exchange & Stock & Best Count \\
        \midrule
        \multicolumn{1}{c}{\multirow{2}{*}{\textbf{ARMD (Ours)}}} & MSE & \textbf{0.167} & \textbf{0.445} & 0.311 & \textbf{0.337} & \textbf{0.181} & \textbf{0.093} & \textbf{0.235} & \multirow{2}{*}{\textbf{12}} \\
          &  MAE & \textbf{0.236} & \textbf{0.459} & \textbf{0.338} & \textbf{0.376} & 0.255 & \textbf{0.203} & \textbf{0.269} & \\
        \midrule
        \multicolumn{1}{c}{\multirow{2}{*}{Diffusion-TS \cite{yuan2024diffusion}}} & MSE & 0.181 & 0.643 & 0.544 & 0.678 & 0.497 & 0.275 & 0.416 & \multirow{2}{*}{0} \\
        & MAE & 0.252 & 0.586 & 0.494 & 0.613 & 0.459 & 0.382 & 0.533 & \\
        \midrule
        \multicolumn{1}{c}{\multirow{2}{*}{MG-TSD  \cite{fan2024mg}}} & MSE & 0.443 & 1.096 & \textbf{0.295} & 0.690 & 0.202 & 0.396 & 0.365 & \multirow{2}{*}{1} \\
        & MAE & 0.529 & 0.765 & 0.345 & 0.631 & 0.278 & 0.460 & 0.453 & \\
        \midrule
        \multicolumn{1}{c}{\multirow{2}{*}{TSDiff \cite{kollovieh2024predict}}} & MSE & 0.352 & 0.614 & 0.470 & 0.686 & 0.242 & 0.125 & 0.330 & \multirow{2}{*}{0} \\
        & MAE & 0.432 & 0.521 & 0.418 & 0.603 & 0.311 & 0.240 & 0.365 & \\
        \midrule
        \multicolumn{1}{c}{\multirow{2}{*}{D3VAE \cite{li2022generative}}} & MSE & 0.416 & 1.123 & 0.389 & 0.644 & 0.394 & 0.240 & 0.345 & \multirow{2}{*}{0} \\
        & MAE & 0.492 & 0.728 & 0.373 & 0.538 & 0.410 & 0.371 & 0.390 & \\
        \midrule
        \multicolumn{1}{c}{\multirow{2}{*}{TimeGrad \cite{timegrad}}} & MSE & 0.359 & 0.884 & 0.297 & 0.661 & 0.182 & 0.508 & 0.333 & \multirow{2}{*}{1} \\
        & MAE & 0.449 & 0.725 & 0.349 & 0.639 & \textbf{0.254} & 0.554 & 0.376 & \\
        \bottomrule
    \end{tabular}}
\vspace{-2mm}
\end{table*}

\begin{table*}[ht]
    \centering
    \setlength{\tabcolsep}{2.0pt}
    \renewcommand{\arraystretch}{1.1}
        \caption{Result comparisons of multivariate series forecasting with other TSF models. The best results are highlighted in bold. The ``Best Count" column indicates the times of achieving the best result.}
        \vspace{-2mm}
        	\scalebox{1.0}{
    \begin{tabular}{p{5.1cm}cccccccccc}
        \toprule
        \multicolumn{1}{c}{Methods} & Metric & Solar Energy & ETTh1 & ETTh2 & ETTm1 & ETTm2 & Exchange & Stock & Best Count \\
        \midrule
        \multicolumn{1}{c}{\multirow{2}{*}{\textbf{ARMD (Ours)}}} & MSE & \textbf{0.167} & 0.445 & 0.311 & 0.337 & 0.181 & 0.093 & \textbf{0.235} & \multirow{2}{*}{\textbf{7}} \\
        & MAE & \textbf{0.236} & 0.459 & \textbf{0.338} & 0.376 & \textbf{0.255} & \textbf{0.203} & \textbf{0.269} & \\
        \midrule
        \multicolumn{1}{c}{\multirow{2}{*}{iTransformer \cite{liu2023itransformer}}} & MSE & 0.203 & 0.386 & \textbf{0.297} & 0.334 & 0.180 & \textbf{0.086} & 0.342 & \multirow{2}{*}{2} \\
        & MAE & 0.237 & 0.405 & 0.349 & 0.368 & 0.264 & 0.206 & 0.413 & \\
        \midrule
        \multicolumn{1}{c}{\multirow{2}{*}{TimesNet \cite{wu2023timesnet}}} & MSE & 0.250 & \textbf{0.384} & 0.340 & 0.338 & 0.187 & 0.107 & 0.427 & \multirow{2}{*}{1} \\
        & MAE & 0.292 & 0.402 & 0.347 & 0.375 & 0.267 & 0.234 & 0.499 & \\
        \midrule
        \multicolumn{1}{c}{\multirow{2}{*}{DLinear \cite{Zeng2022AreTE}}} & MSE & 0.290 & 0.386 & 0.333 & 0.345 & 0.193 & 0.088 & 0.286 & \multirow{2}{*}{1} \\
        & MAE & 0.378 & \textbf{0.400} & 0.387 & 0.372 & 0.292 & 0.218 & 0.325 & \\
        \midrule
        \multicolumn{1}{c}{\multirow{2}{*}{PatchTST \cite{nie2022time}}} & MSE & 0.234 & 0.414 & 0.302 & \textbf{0.329} & \textbf{0.175} & 0.088 & 0.516 & \multirow{2}{*}{3} \\
        & MAE & 0.286 & 0.419 & 0.348 & \textbf{0.367} & 0.259 & 0.205 & 0.524 & \\
        \midrule
        \multicolumn{1}{c}{\multirow{2}{*}{Client \cite{gao2023client}}} & MSE & 0.199 & 0.392 & 0.305 & 0.336 & 0.184 & \textbf{0.086} & 0.352 & \multirow{2}{*}{1} \\
        & MAE & 0.239 & 0.409 & 0.353 & 0.369 & 0.267 & 0.206 & 0.433 & \\
        \bottomrule
    \end{tabular}}
\vspace{-3mm}
    \label{tab:main-multi1}
\end{table*}

\subsection{Experimental Settings}

The proposed ARMD is extensively evaluated on seven widely used benchmark datasets, including Solar Energy \cite{lai2018modeling}, Exchange \cite{lai2018modeling}, Stock \cite{yoon2019time}, and four ETT datasets \cite{haoyietal-informer-2021}. We compare ARMD with five advanced diffusion-based TSF models: Diffusion-TS \cite{yuan2024diffusion}, MG-TSD \cite{fan2024mg}, TSDiff \cite{kollovieh2024predict}, D3VAE \cite{li2022generative}, TimeGrad \cite{timegrad}. 
Additionally, some other advanced TSF models, including iTransformer \cite{liu2023itransformer}, TimesNet \cite{wu2023timesnet}, DLinear \cite{Zeng2022AreTE}, PatchTST \cite{nie2022time}, and Client \cite{gao2023client} are also compared with ARMD.

For all datasets, the historical length and prediction length are both set to 96. Following the evaluation methodology employed in a previous study \cite{haoyietal-informer-2021}, we calculate the mean squared error (MSE) and mean absolute error (MAE) on z-score normalized data, enabling a consistent assessment of various variables. More details concerning experiment settings are shown in the supplemental materials.

\subsection{Comparison with Diffusion-Based Models}
The multivariate forecasting results, comparing our proposed ARMD with various advanced diffusion-based TSF models, are summarized in Table \ref{tab:main-multi}, with each model's results averaged over 10 sampling runs. Notably, ARMD demonstrates superior forecasting performance, achieving optimal results in 12 out of the 14 experimental settings. Our extensive experiments reveal that while other diffusion-based TSF models can achieve effective results on specific datasets, they often suffer from instability, performing well in certain scenarios but deviating significantly from the true values in others, which indicates a lack of generalization across diverse data. In contrast, ARMD consistently exhibits robust performance across all datasets, suggesting a higher degree of reliability and adaptability to various multivariate forecasting challenges. Specifically, on the ETTm1 dataset, ARMD achieves a substantial 47.7\% reduction in MSE and a 30.1\% reduction in MAE compared to the second-best model, D3VAE. On the Stock dataset, ARMD surpasses TSDiff, the second-best model in this setting, with a 28.8\% reduction in MSE and a 26.3\% reduction in MAE. Furthermore, ARMD attains more than a 10\% reduction in both MSE and MAE on the Solar Energy, ETTh1, and Exchange datasets compared to the nearest competitor. These results indicate that ARMD is the best diffusion-based TSF model.

\begin{algorithm}[!t]
\caption{Sampling/Forecasting.}\label{alg:sampling}
\begin{algorithmic}[1]
\REQUIRE Historical series ${X^{T}_{-T+1:0}}$; Trained devolution network $R(.)$; Sampling interval $\Delta t$; Predefined coefficients $\bar{\alpha}_{0:T}$. 
\FOR{$t = T$ \textbf{to} $0$ \textbf{by} $\Delta t$}
    \STATE Obtain $\hat{X^{0}}(X^{t},t,\theta)$ using ${X^{t}_{1-t:T-t}}$ and $t$ with the devolution network $R(.)$, and calculate the corresponding evolution trend $\hat{z}(t, \theta)$ using Equation (\ref{eq:zt}); 
    \STATE Update $X^{t}_{1-t:T-t}$ using Equation (\ref{equ:reverse_k});
\ENDFOR
\STATE Output the prediction of $X^{0}_{1:T}$.
\end{algorithmic}
\end{algorithm}

\subsection{Comparison with Other TSF Models}

As demonstrated in Table \ref{tab:main-multi1}, ARMD consistently outperforms other advanced models in multivariate TSF tasks. The comparative results for other TSF models are primarily sourced from iTransformer \cite{liu2023itransformer}, with the Stock dataset being an exception, where additional evaluations are conducted using official implementations to ensure a thorough analysis. ARMD achieves the highest number of best counts, indicating its consistent ability to surpass other TSF methods across multiple benchmarks. These results not only establish ARMD as the leading diffusion-based TSF model but also underscore its superiority over existing models. Overall, these findings highlight ARMD's promise as a robust and effective solution for multivariate TSF tasks.



\subsection{Qualitative Analysis}
\begin{figure}[!htbp]
\centering
\vspace{-3mm}
\includegraphics[width=0.75\linewidth]{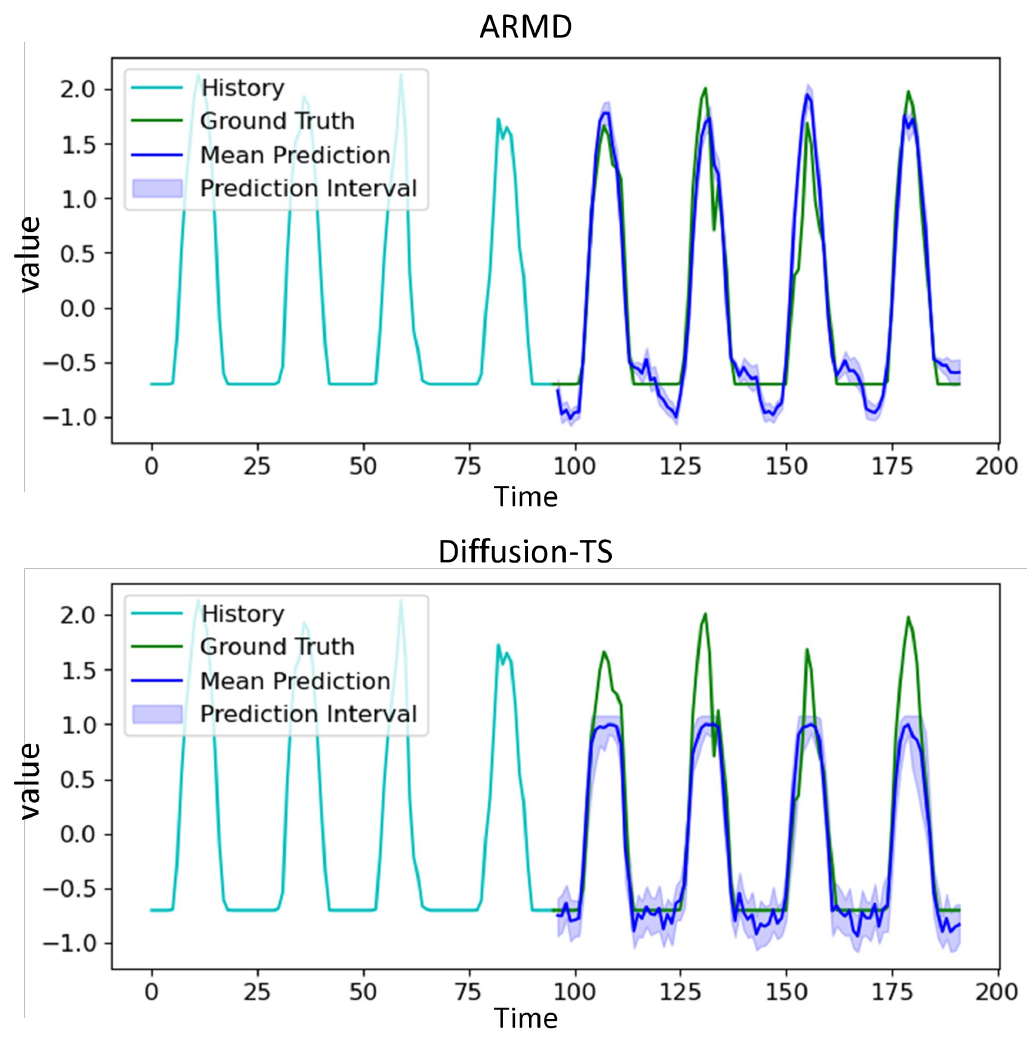} 
\vspace{-3mm}
\caption{The distributions of 10 different predictions made by ARMD and Diffusion-TS given the same historical series. ARMD achieves more stable and accurate predictions.}
\vspace{-6mm}
\label{fig:uncertain}
\end{figure}

As illustrated in Fig. \ref{fig:uncertain}, we qualitatively analyze the performance of ARMD by comparing it with the advanced diffusion-based model, Diffusion-TS. Given the same historical series, both models make 10 independent predictions for the future series. Our analysis reveals that ARMD consistently produces more stable and accurate predictions, particularly in scenarios involving high peak values. This is in contrast to Diffusion-TS, which shows greater variability and less precision in similar settings. We attribute ARMD's superior performance to its deterministic training process and the reduced sampling steps required during inference. These features of ARMD not only enhance the model’s ability to generate predictions that closely align with actual series but also result in narrower uncertainty distributions. 

\begin{table}[!t]
\centering
\vspace{-1mm}
\caption{Training/inference time (s) of ARMD, Diffusion-TS, MG-TSD and TimeGrad on the ETTm1 dataset, and the historical and prediction length are both set to 96.}
\vspace{-2mm}
\label{tab:time}
\begin{tabular}{ccc}
\toprule
 & Training time & Inference time\\
\midrule
\textbf{ARMD (Ours)}  & \textbf{11.872} & \textbf{31.650} \\
\midrule
Diffusion-TS  & 149.974 & 1183.593 \\
\midrule
MG-TSD  & 834.815 & 3299.918 \\
\midrule
TimeGrad  & 449.792 & 1749.896 \\
\bottomrule
\end{tabular}
\vspace{-5mm}
\end{table}

\begin{table*}[!t]
    \centering
        \caption{Ablation experimental results. \textbf{Interpolation Method}: replacing the intermediate state generation method with an interpolation method. \textbf{T-embedding Method}: using a $t$-embedding approach for denoising. \textbf{Transformer Backbone}: utilizing a Transformer-based backbone.  
        \textbf{Removing Deviation}: removing deviations during the training process.  \textbf{Adding Noise}: adding sampling noise. The best results are highlighted in bold.}
    \setlength{\tabcolsep}{4pt}
    \renewcommand{\arraystretch}{1.1}
    \begin{tabular}{lcccccccccc}
        \toprule
        \multicolumn{1}{c}{Methods}& Metric & Solar Energy & ETTh1 & ETTh2 & ETTm1 & ETTm2 & Exchange & Stock & Best Count \\
        \midrule
        \multicolumn{1}{c}{\multirow{2}{*}{\textbf{ARMD (Ours)}}} & MSE & \textbf{0.167} & \textbf{0.445} & 0.311 & \textbf{0.337} & 0.181 & \textbf{0.093} & \textbf{0.235} & \multirow{2}{*}{\textbf{11}} \\    
        & MAE & \textbf{0.236} & \textbf{0.459} & \textbf{0.338} & \textbf{0.376} & 0.255 & \textbf{0.203} & \textbf{0.269} & \\
        \midrule
         \multicolumn{1}{c}{\multirow{2}{*}{Interpolation Method}} & MSE & 0.184 & 0.509 & 0.407 & 0.359 & 0.210 & 0.131 & 0.249 & \multirow{2}{*}{0} \\
        & MAE & 0.250 & 0.481 & 0.373 & 0.385 & 0.265 & 0.249 & 0.278 & \\
        \midrule
         \multicolumn{1}{c}{\multirow{2}{*}{T-embedding Method}} & MSE & 0.308 & 0.707 & 0.376 & 0.598 & 0.235 & 0.233 & 0.350 & \multirow{2}{*}{0} \\
        & MAE & 0.380 & 0.589 & 0.380 & 0.527 & 0.331 & 0.330 & 0.423 & \\
        \midrule
         \multicolumn{1}{c}{\multirow{2}{*}{Transformer Backbone}} & MSE & 1.343 & 0.631 & \textbf{0.285} & 0.578 & \textbf{0.168} & 0.137 & 0.239 & \multirow{2}{*}{3} \\
        & MAE & 0.713 & 0.531 & 0.341 & 0.499 & \textbf{0.253} & 0.242 & 0.271 & \\
        \midrule
         \multicolumn{1}{c}{\multirow{2}{*}{Removing Deviation}} & MSE & 0.183 & 0.473 & 0.368 & 0.375 & 0.211 & 0.133 & 0.248 & \multirow{2}{*}{0}  \\
       & MAE & 0.254 & 0.467 & 0.358 & 0.391 & 0.266 & 0.250 & 0.277 &  \\
        \midrule
         \multicolumn{1}{c}{\multirow{2}{*}{Adding Noise}} & MSE & 0.199 & 0.482 & 0.334 & 0.552 & 0.199 & 0.103 & 0.328 & \multirow{2}{*}{0}  \\
       & MAE & 0.315 & 0.479 & 0.353 & 0.526 & 0.295 & 0.228 & 0.347 &  \\
        \bottomrule
    \end{tabular}
    \label{tab:abla1}
\vspace{-1.5mm}
\end{table*}

\subsection{Efficiency Comparison}
We compare the training and inference efficiency of ARMD with the diffusion-based TSF models Diffusion-TS, MG-TSD, and TimeGrad on the ETTm1 dataset, as presented in Table \ref{tab:time}. Due to the optimized model structure and the reduction in the number of sampling steps, ARMD achieves significantly shorter training and inference time compared to other diffusion models, resulting in more than a tenfold acceleration in both training and inference.

\subsection{Ablation Studies}

In this section, we set a range of ablation experiments to testify diverse components of the proposed ARMD.

\noindent \textbf{Intermediate State Generation Method.} 
ARMD generates intermediate states using a sliding-based method to maintain series continuity. An alternative method to generate intermediate states involves interpolating between the initial series ${X^{0}_{1:T}}$ and final series ${X^{T}_{-T+1:0}}$, as described by: 
\[
\begin{aligned}
X^{t}_{1-t:T-t} = X^{0}_{1:T} + (X^{T}_{-T+1:0}-X^{0}_{1:T})*t/T .
\label{eq:diff2t_0}
\end{aligned}
\]
The results of using these interpolation-based intermediate states are presented in the second row of Table \ref{tab:abla1}. The results indicate that our sliding-based method consistently outperforms the interpolation-based approach across all settings, underscoring its superiority in preserving series continuity.

\noindent \textbf{Devolution/Denosing Learning Method.} The devolution network of ARMD take a distance-based method to provide the prediction $\hat{X^{0}}(X^{t},t,\theta)$ and $\hat{z}(t, \theta)$, differing from conventional approaches. Traditional denoising (devolution) networks often utilize a $t$-embedding-based method \cite{yuan2024diffusion, shen2023non, shen2024multi}. In this approach, the model initially generates a prediction based on the input intermediate state, then embeds the time step $t$ to generate a $t$-embedding, and finally combines the $t$-embedding with the initial prediction to produce the final output. The results of applying this $t$-embedding-based method within the devolution network are shown in the third row of Table \ref{tab:abla1}, demonstrating that our distance-based method outperforms the $t$-embedding-based method across all settings.

\noindent \textbf{Backbone of the Devolution Network.} Many existing diffusion-based TSF models use a Transformer-based backbone in the denoising network \cite{yuan2024diffusion, litransformer, tashiro2021csdi}. Nevertheless, the devolution network in ARMD uses a linear-based backbone, which accelerates training and sampling. The impact of replacing the linear-based backbone with a Transformer-based backbone is presented in the fourth row of Table \ref{tab:abla1}. The results show that the linear-based backbone outperforms the Transformer-based backbone in most settings (11 out of 14).


\noindent \textbf{Deviations to the Intermediate States.} To enhance the diversity of intermediate states and prevent over-fitting, minor deviations are added to the input of the devolution network during the training process. The consequences of removing these deviations are detailed in the fifth row of Table \ref{tab:abla1}, highlighting their critical role in preventing over-fitting.

\noindent \textbf{Noise in the Sampling Process.} Given that the series sampling/forecasting process is deterministic, the noise term $\sigma_t \epsilon_t$ in Equation (\ref{equ:reverse_origin}) is set to 0 during sampling/forecasting. The results of adding sampling noise are shown in the last row of Table \ref{tab:abla1}, showing that introducing randomness leads to a decline in model performance.



\section{Conclusion}
Inspired by the classic ARMA theory, we creatively introduce an Auto-Regressive Moving Diffusion (ARMD) model for TSF, which reinterprets the evolution of time series as a diffusion process. Particularly, regarding the target future series as the initial state, and the historical series as the final state, the intermediate series are generated with the series slide operations to complete the diffusion process. In this manner, the sampling procedure of the diffusion model becomes the series forecasting, which denotes that the diffusion mechanism is successfully aligned with the TSF objective, and further resulting in an unconditional, continuous sequential diffusion TSF model. The proposed ARMD is validated across seven widely used datasets, and the experimental results show that our method effectively suits the unique characteristics of time series data, and achieves superior forecasting performance.

\section{Acknowledgements}
This work is financially supported by the The Major Science and Technology Projects of Ningbo (No. 2022Z236), Natural Science Foundation of Ningbo of China (No. 2023J027), China Meteorological Administration under Grant QBZ202316 as well as by the High Performance Computing Centers at Eastern Institute of Technology, Ningbo, and Ningbo Institute of Digital Twin. We would like to express our gratitude to Associate Professor Wenbo Hu from Hefei University of Technology, Associate Professor Hao Sun from Renmin University of China, and Professor Cewu Lu from Shanghai Jiao Tong University for their assistance with this work.

\bibliography{aaai25}

\begin{thebibliography}{40}
\providecommand{\natexlab}[1]{#1}

\bibitem[{Alcaraz and Strodthoff(2022)}]{alcaraz2022diffusion}
Alcaraz, J. M.~L.; and Strodthoff, N. 2022.
\newblock Diffusion-based time series imputation and forecasting with structured state space models.
\newblock \emph{arXiv preprint arXiv:2208.09399}.

\bibitem[{Benny and Wolf(2022)}]{benny2022dynamic}
Benny, Y.; and Wolf, L. 2022.
\newblock Dynamic dual-output diffusion models.
\newblock In \emph{Computer Vision and Pattern Recognition}.

\bibitem[{Box et~al.(2015)Box, Jenkins, Reinsel, and Ljung}]{box2015time}
Box, G.~E.; Jenkins, G.~M.; Reinsel, G.~C.; and Ljung, G.~M. 2015.
\newblock \emph{Time series analysis: forecasting and control}.
\newblock John Wiley \& Sons.

\bibitem[{Cao et~al.(2024)Cao, Wang, Li, Chen, Ma, and Yang}]{cao2024teachingvideodiffusionmodel}
Cao, Q.; Wang, D.; Li, X.; Chen, Y.; Ma, C.; and Yang, X. 2024.
\newblock Teaching Video Diffusion Model with Latent Physical Phenomenon Knowledge.
\newblock arXiv:2411.11343.

\bibitem[{Chen et~al.(2023)Chen, Sun, Song, and Luo}]{chen2023diffusiondet}
Chen, S.; Sun, P.; Song, Y.; and Luo, P. 2023.
\newblock Diffusiondet: Diffusion model for object detection.
\newblock In \emph{Proceedings of the IEEE/CVF international conference on computer vision}, 19830--19843.

\bibitem[{Croitoru et~al.(2023)Croitoru, Hondru, Ionescu, and Shah}]{croitoru2023diffusion}
Croitoru, F.-A.; Hondru, V.; Ionescu, R.~T.; and Shah, M. 2023.
\newblock Diffusion models in vision: A survey.
\newblock \emph{IEEE Transactions on Pattern Analysis and Machine Intelligence}, 45(9): 10850--10869.

\bibitem[{Fan et~al.(2024)Fan, Wu, Xu, Huang, Liu, and Bian}]{fan2024mg}
Fan, X.; Wu, Y.; Xu, C.; Huang, Y.; Liu, W.; and Bian, J. 2024.
\newblock MG-TSD: Multi-granularity time series diffusion models with guided learning process.
\newblock \emph{arXiv preprint arXiv:2403.05751}.

\bibitem[{Fang et~al.(2022)Fang, Zhuo, Yan, Song, Jiang, and Zhou}]{fang2022attention}
Fang, W.; Zhuo, W.; Yan, J.; Song, Y.; Jiang, D.; and Zhou, T. 2022.
\newblock Attention meets long short-term memory: A deep learning network for traffic flow forecasting.
\newblock \emph{Physica A: Statistical Mechanics and its Applications}, 587: 126485.

\bibitem[{Gao et~al.(2024)Gao, Cao, Chen, and Zhang}]{gao2024crossvariablelinearintegratedenhanced}
Gao, J.; Cao, Q.; Chen, Y.; and Zhang, D. 2024.
\newblock Cross-variable Linear Integrated ENhanced Transformer for Photovoltaic power forecasting.
\newblock arXiv:2406.03808.

\bibitem[{Gao et~al.(2023{\natexlab{a}})Gao, Chen, Hu, and Zhang}]{gao2023adaptive}
Gao, J.; Chen, Y.; Hu, W.; and Zhang, D. 2023{\natexlab{a}}.
\newblock An adaptive deep-learning load forecasting framework by integrating Transformer and domain knowledge.
\newblock \emph{Advances in Applied Energy}, 100142.

\bibitem[{Gao, Hu, and Chen(2023)}]{gao2023client}
Gao, J.; Hu, W.; and Chen, Y. 2023.
\newblock Client: Cross-variable linear integrated enhanced transformer for multivariate long-term time series forecasting.
\newblock \emph{arXiv preprint arXiv:2305.18838}.

\bibitem[{Gao et~al.(2023{\natexlab{b}})Gao, Hu, Cao, Dai, and Chen}]{gao2023clearforecastcontrastivelearninghighpurity}
Gao, J.; Hu, Y.; Cao, Q.; Dai, S.; and Chen, Y. 2023{\natexlab{b}}.
\newblock CLeaRForecast: Contrastive Learning of High-Purity Representations for Time Series Forecasting.
\newblock arXiv:2312.05758.

\bibitem[{Ho, Jain, and Abbeel(2020)}]{ddpm20}
Ho, J.; Jain, A.; and Abbeel, P. 2020.
\newblock Denoising diffusion probabilistic models.
\newblock In \emph{Neural Information Processing Systems}.

\bibitem[{Kollovieh et~al.(2024)Kollovieh, Ansari, Bohlke-Schneider, Zschiegner, Wang, and Wang}]{kollovieh2024predict}
Kollovieh, M.; Ansari, A.~F.; Bohlke-Schneider, M.; Zschiegner, J.; Wang, H.; and Wang, Y.~B. 2024.
\newblock Predict, refine, synthesize: Self-guiding diffusion models for probabilistic time series forecasting.
\newblock \emph{Advances in Neural Information Processing Systems}, 36.

\bibitem[{Kong et~al.(2020)Kong, Ping, Huang, Zhao, and Catanzaro}]{kong2020diffwave}
Kong, Z.; Ping, W.; Huang, J.; Zhao, K.; and Catanzaro, B. 2020.
\newblock Diffwave: A versatile diffusion model for audio synthesis.
\newblock \emph{arXiv preprint arXiv:2009.09761}.

\bibitem[{Lai et~al.(2018)Lai, Chang, Yang, and Liu}]{lai2018modeling}
Lai, G.; Chang, W.-C.; Yang, Y.; and Liu, H. 2018.
\newblock Modeling long-and short-term temporal patterns with deep neural networks.
\newblock In \emph{The 41st international ACM SIGIR conference on research \& development in information retrieval}, 95--104.

\bibitem[{Li et~al.(2024)Li, Chen, Hu, Chen, and Zhou}]{litransformer}
Li, Y.; Chen, W.; Hu, X.; Chen, B.; and Zhou, M. 2024.
\newblock Transformer-Modulated Diffusion Models for Probabilistic Multivariate Time Series Forecasting.
\newblock In \emph{The Twelfth International Conference on Learning Representations}.

\bibitem[{Li et~al.(2022)Li, Lu, Wang, and Dou}]{li2022generative}
Li, Y.; Lu, X.; Wang, Y.; and Dou, D. 2022.
\newblock Generative time series forecasting with diffusion, denoise, and disentanglement.
\newblock \emph{Advances in Neural Information Processing Systems}, 35: 23009--23022.

\bibitem[{Liu et~al.(2024)Liu, Hu, Zhang, Wu, Wang, Ma, and Long}]{liu2023itransformer}
Liu, Y.; Hu, T.; Zhang, H.; Wu, H.; Wang, S.; Ma, L.; and Long, M. 2024.
\newblock iTransformer: Inverted Transformers Are Effective for Time Series Forecasting.
\newblock In \emph{The Twelfth International Conference on Learning Representations}.

\bibitem[{Lopez-Lira and Tang(2023)}]{lopez2023can}
Lopez-Lira, A.; and Tang, Y. 2023.
\newblock Can ChatGPT Forecast Stock Price Movements? Return Predictability and Large Language Models.
\newblock \emph{arXiv preprint arXiv:2304.07619}.

\bibitem[{Lovelace et~al.(2024)Lovelace, Kishore, Wan, Shekhtman, and Weinberger}]{lovelace2024latent}
Lovelace, J.; Kishore, V.; Wan, C.; Shekhtman, E.; and Weinberger, K.~Q. 2024.
\newblock Latent diffusion for language generation.
\newblock \emph{Advances in Neural Information Processing Systems}, 36.

\bibitem[{Lugmayr et~al.(2022)Lugmayr, Danelljan, Romero, Yu, Timofte, and Van~Gool}]{lugmayr2022repaint}
Lugmayr, A.; Danelljan, M.; Romero, A.; Yu, F.; Timofte, R.; and Van~Gool, L. 2022.
\newblock Repaint: Inpainting using denoising diffusion probabilistic models.
\newblock In \emph{IEEE/CVF Conference on Computer Vision and Pattern Recognition}.

\bibitem[{Meijer and Chen(2024)}]{meijer2024rise}
Meijer, C.; and Chen, L.~Y. 2024.
\newblock The Rise of Diffusion Models in Time-Series Forecasting.
\newblock \emph{arXiv preprint arXiv:2401.03006}.

\bibitem[{Nie et~al.(2022)Nie, Nguyen, Sinthong, and Kalagnanam}]{nie2022time}
Nie, Y.; Nguyen, N.~H.; Sinthong, P.; and Kalagnanam, J. 2022.
\newblock A Time Series is Worth 64 Words: Long-term Forecasting with Transformers.
\newblock \emph{arXiv preprint arXiv:2211.14730}.

\bibitem[{Piao et~al.(2024)Piao, Chen, Murayama, Matsubara, and Sakurai}]{piao2024fredformer}
Piao, X.; Chen, Z.; Murayama, T.; Matsubara, Y.; and Sakurai, Y. 2024.
\newblock Fredformer: Frequency Debiased Transformer for Time Series Forecasting.
\newblock \emph{arXiv preprint arXiv:2406.09009}.

\bibitem[{Rasul et~al.(2021)Rasul, Seward, Schuster, and Vollgraf}]{timegrad}
Rasul, K.; Seward, C.; Schuster, I.; and Vollgraf, R. 2021.
\newblock Autoregressive denoising diffusion models for multivariate probabilistic time series forecasting.
\newblock In \emph{International Conference on Machine Learning}.

\bibitem[{Rombach et~al.(2022)Rombach, Blattmann, Lorenz, Esser, and Ommer}]{rombach2022high}
Rombach, R.; Blattmann, A.; Lorenz, D.; Esser, P.; and Ommer, B. 2022.
\newblock High-resolution image synthesis with latent diffusion models.
\newblock In \emph{Proceedings of the IEEE/CVF conference on computer vision and pattern recognition}, 10684--10695.

\bibitem[{Sen, Yu, and Dhillon(2019)}]{Sen2019ThinkGA}
Sen, R.; Yu, H.-F.; and Dhillon, I.~S. 2019.
\newblock Think globally, act locally: A deep neural network approach to high-dimensional time series forecasting.
\newblock \emph{Advances in neural information processing systems}, 32.

\bibitem[{Shen, Chen, and Kwok(2024)}]{shen2024multi}
Shen, L.; Chen, W.; and Kwok, J. 2024.
\newblock Multi-Resolution Diffusion Models for Time Series Forecasting.
\newblock In \emph{The Twelfth International Conference on Learning Representations}.

\bibitem[{Shen and Kwok(2023)}]{shen2023non}
Shen, L.; and Kwok, J. 2023.
\newblock Non-autoregressive conditional diffusion models for time series prediction.
\newblock In \emph{International Conference on Machine Learning}, 31016--31029. PMLR.

\bibitem[{Song, Meng, and Ermon(2020)}]{song2020denoising}
Song, J.; Meng, C.; and Ermon, S. 2020.
\newblock Denoising diffusion implicit models.
\newblock \emph{arXiv preprint arXiv:2010.02502}.

\bibitem[{Tashiro et~al.(2021)Tashiro, Song, Song, and Ermon}]{tashiro2021csdi}
Tashiro, Y.; Song, J.; Song, Y.; and Ermon, S. 2021.
\newblock {CSDI}: Conditional score-based diffusion models for probabilistic time series imputation.
\newblock In \emph{Neural Information Processing Systems}.

\bibitem[{Wen et~al.(2022)Wen, Zhou, Zhang, Chen, Ma, Yan, and Sun}]{wen2022transformers}
Wen, Q.; Zhou, T.; Zhang, C.; Chen, W.; Ma, Z.; Yan, J.; and Sun, L. 2022.
\newblock Transformers in time series: A survey.
\newblock \emph{arXiv preprint arXiv:2202.07125}.

\bibitem[{Wu et~al.(2023)Wu, Hu, Liu, Zhou, Wang, and Long}]{wu2023timesnet}
Wu, H.; Hu, T.; Liu, Y.; Zhou, H.; Wang, J.; and Long, M. 2023.
\newblock TimesNet: Temporal 2D-Variation Modeling for General Time Series Analysis.
\newblock In \emph{International Conference on Learning Representations}.

\bibitem[{Xu et~al.(2024)Xu, Wu, Li, Danoy, and Bouvry}]{xu2024survey}
Xu, J.; Wu, C.; Li, Y.-F.; Danoy, G.; and Bouvry, P. 2024.
\newblock Survey and Taxonomy: The Role of Data-Centric AI in Transformer-Based Time Series Forecasting.
\newblock \emph{arXiv preprint arXiv:2407.19784}.

\bibitem[{Yang et~al.(2024)Yang, Jin, Wen, Zhang, Liang, Ma, Wang, Liu, Yang, Xu et~al.}]{yang2024survey}
Yang, Y.; Jin, M.; Wen, H.; Zhang, C.; Liang, Y.; Ma, L.; Wang, Y.; Liu, C.; Yang, B.; Xu, Z.; et~al. 2024.
\newblock A survey on diffusion models for time series and spatio-temporal data.
\newblock \emph{arXiv preprint arXiv:2404.18886}.

\bibitem[{Yoon, Jarrett, and Van~der Schaar(2019)}]{yoon2019time}
Yoon, J.; Jarrett, D.; and Van~der Schaar, M. 2019.
\newblock Time-series generative adversarial networks.
\newblock \emph{Advances in neural information processing systems}, 32.

\bibitem[{Yuan and Qiao(2024)}]{yuan2024diffusion}
Yuan, X.; and Qiao, Y. 2024.
\newblock Diffusion-ts: Interpretable diffusion for general time series generation.
\newblock \emph{arXiv preprint arXiv:2403.01742}.

\bibitem[{Zeng et~al.(2023)Zeng, Chen, Zhang, and Xu}]{Zeng2022AreTE}
Zeng, A.; Chen, M.; Zhang, L.; and Xu, Q. 2023.
\newblock Are transformers effective for time series forecasting?
\newblock In \emph{Proceedings of the AAAI conference on artificial intelligence}, volume~37, 11121--11128.

\bibitem[{Zhou et~al.(2021)Zhou, Zhang, Peng, Zhang, Li, Xiong, and Zhang}]{haoyietal-informer-2021}
Zhou, H.; Zhang, S.; Peng, J.; Zhang, S.; Li, J.; Xiong, H.; and Zhang, W. 2021.
\newblock Informer: Beyond efficient transformer for long sequence time-series forecasting.
\newblock In \emph{Proceedings of the AAAI conference on artificial intelligence}, volume~35, 11106--11115.

\end{thebibliography}

\clearpage
\appendix
\begin{center}
\huge Supplemental Materials for ``Auto-Regressive Moving Diffusion Models for Time Series Forecasting"
\end{center}

\section{Preliminary of Diffusion Model and Conditional DDPMs for TSF} \label{app:diffu}
Diffusion models have been widely applied in various fields \cite{rombach2022high, kong2020diffwave, lovelace2024latent, chen2023diffusiondet}, typically comprising a forward diffusion process and a reverse process. A prominent example of these models is the Denoising Diffusion Probabilistic Model (DDPM) \cite{ddpm20}. In the forward diffusion process of DDPM, noise is gradually added to the input $X^0$, transforming it into a white Gaussian noise $X^T$ over $T$ diffusion steps. At each time step $t\in[1,T]$, the diffused sample $X^t$ is acquired through scaling the previous sample $X^{t-1}$ by $\sqrt{1-\beta_t}$ and subsequently introducing noise, as described by:
\begin{equation}
  q(X^t|X^{t-1})=\mathcal{N}(X^t; \sqrt{1-\beta_t}X^{t-1},\beta_tI),
\end{equation}
Here, $\beta_t\in[0, 1]$ denotes the noise variance, which follows a predefined schedule. From this, it can be derived that:
\begin{equation}
q(X^t|X^{0})=\mathcal{N}(X^t;\sqrt{\bar{\alpha}_t}X^{0},(1-\bar{\alpha}_t)I).
\label{eq:x0_2_xt}
\end{equation}
where $\alpha_t=1-\beta_t$ and:  
\begin{equation}
\bar{\alpha}_t=\Pi_{k=1}^t\alpha_k. 
\label{eq:alpha}
\end{equation}
As $t$ increases, $\bar{\alpha}_t$ decreases, and it is chosen as predefined deviation coefficient and devolution coefficient in our proposed ARMD. In DDPM, $X^t$ can be directly computed as:
\begin{align}
X^t=\sqrt{\bar{\alpha}_t}X^{0}+\sqrt{1-\bar{\alpha}_t}\epsilon,
\label{eq:x0_samples_xt}
\end{align}
where $\epsilon$ is sampled from $\mathcal{N}(\mathrm{0}, I)$. 
The reverse denoising process of DDPM is a Markovian process, expressed as:
\begin{equation}
  p_{\theta}(X^{t-1}|X^t)=\mathcal{N}(X^{t-1};\mu_{\theta}(X^t,t),\Sigma_{\theta}(X^t,t)).
\label{eq:denoise_t}
\end{equation}
In practice, $\Sigma_{\theta}(X^t,t)$ is generally kept constant at $\sigma_t^2I$, while $\mu_{\theta}(X^t,t)$ is usually predicted by a neural network parameterized by $\theta$. ${\mu}_{\theta}(X^t,t)$ can be defined in two ways \cite{benny2022dynamic}: (1) noise-based ${\mu}_{\epsilon}(\epsilon_{\theta})$ or (2) data-based ${\mu}_{X}(X_{\theta})$, and these two different definitions can lead to different optimization objectives.

In TSF, the aim is to forecast the future values ${X_{1:F}}$ of a time series given its historical series ${X_{-L+1:0}}$. Here, $F$ is the length of the series to be predicted, and $L$ is the length of the historical series. Previous approaches typically utilize conditional DDPMs for TSF, modeling the distribution:
\begin{equation}
  p_{\theta}(X_{1:F}^{0:T}|c)=p_{\theta}(X_{1:F}^{T})\prod_{t=1}^Tp_{\theta}(X_{1:F}^{t-1}|X_{1:F}^{t},c),
\end{equation}
where $X_{1:F}^{T}\sim\mathcal{N}(\mathbf{0},I)$, and $c=g(X_{-L+1:0}^0)$ is the output of the conditioning network $g$ that takes the historical series ${X_{-L+1:0}}$ as input, and:			
\begin{equation}
p_{\theta}(X_{1:F}^{t-1}|X_{1:F}^{t}, c)
=\mathcal{N}(X_{1:F}^{t-1};\mu_{\theta}(X_{1:F}^{t},t|c),\sigma_t^2I).    
\end{equation}

However, this conditional-based approach is less straightforward compared to the method employed by ARMD.

\section{Connection between ARMD and ARMA}
The Autoregressive Moving Average (ARMA) model is a classic approach in TSF that combines two components:

\noindent \textbf{Autoregressive (AR) Component:} The AR component models the time series using past data points. It assumes that the current data point can be expressed as a linear combination of previous points and a noise term. 
\begin{equation}
x_t = \phi_1 x_{t-1} + \phi_2 x_{t-2} + \dots + \phi_p x_{t-p} + \epsilon_t.
\end{equation}
AR components are particularly adept at handling time series data with long-term trends.

\noindent \textbf{Moving Average (MA) Component:} The MA component models the time series using past forecast errors (noises). It assumes that the current data point can be expressed as a linear combination of past error (noise) terms.
\begin{equation}
x_t = \mu + \theta_1 \epsilon_{t-1} + \theta_2 \epsilon_{t-2} + \dots + \theta_q \epsilon_{t-q} + \epsilon_t.
\end{equation}
MA components can effectively handle time series data with sudden changes or significant noise.

\noindent \textbf{ARMA Model:} Combining both components, the ARMA model can be represented as:
\begin{equation}
\begin{aligned}
x_t = \phi_1 x_{t-1} + \phi_2 x_{t-2} + \dots + \phi_p x_{t-p}  \\ + \theta_1 \epsilon_{t-1} + \theta_2 \epsilon_{t-2} + \dots + \theta_q \epsilon_{t-q} + \epsilon_t.
\end{aligned}
\label{arma}
\end{equation}
The ARMA model combines the strengths of AR and MA components, capturing trend information in time series while also accommodating sudden changes.

In our proposed ARMD, instead of explicitly adding noise, the forward diffusion (evolution) process in our model involves sliding the time series data from the future series ${X^{0}_{1:T}}$ to the historical series ${X^{T}_{-T+1:0}}$. For each diffusion step from ${X^{t}_{1-t:T-t}}$ to ${X^{t+k}_{1-t-k:T-t-k}}$, the process can be viewed as incorporating noise into the series based on the ARMA assumption, and each data point $x_i$ in the shifted series ${X^{t+k}_{1-t-k:T-t-k}}$ contains the introduced noises from the time steps of $i+1:i+k$.

During the reverse denoising (devolution) process, the model devolves ${X^{t+k}_{1-t-k:T-t-k}}$ to ${X^{t}_{1-t:T-t}}$. The devolution network of ARMD is linear-based, wherein each data point $x_i$ can be modeled as a linear combination of the preceding $k$ time steps. This also aligns with the ARMA assumption. These consistencies with ARMA provide theoretical support for the effectiveness of our diffusion model, ARMD, in time series forecasting.

\section{Deviation added in the Training Process} \label{app:deviation}
A small proportion of deviation is added to the input of the devolution network to increase sample diversity and prevent over-fitting during the training process. Specifically, for the input $X^{t}_{1-t:T-t}$, the corresponding deviation can be expressed as $\eta_{0:t}*\epsilon$, where ${\eta}_{0:t}$ is the deviation coefficient following a predefined schedule, and $\epsilon$ is sampled from $\mathcal{N}(\mathbf{0}, \mathbf{I})$. Here, we utilize the predefined diffusion coefficients in DDPM, where ${\eta}_{0:t}$ equals ${\bar{\alpha}_t}$ in equation (\ref{eq:alpha}), with the deviation proportion being smaller when closer to the historical series and larger when closer to the future series. This choice is reasonable because the closer $X^{t}_{1-t:T-t}$ is to the historical series ${X^{T}_{-T+1:0}}$, the more it aligns with the ultimate goal of predicting ${X^{0}_{1:T}}$ from ${X^{T}_{-T+1:0}}$, thus the deviation should be correspondingly reduced.

\section{Hyper-parameters of the Devolution Network} \label{app:hyperparameters_bcd}
The hyper-parameters $b$, $c$, and $d$ in Equation (\ref{equ:reverse_0}) are utilized to balance the relation between the distance prediction $D$ and the input ${X^{t}_{1-t:T-t}}$ of the devolution network $R(.)$. These hyper-parameters vary across different datasets. In our experiments, the hyper-parameters $b$, $c$, and $d$ are selected through a grid search. Specifically, $b$ is chosen from \{1, 1.5, 2\}, $c$ is chosen from \{-1, -0.5, 0.5, 1\}, and $d$ is chosen from \{0.3, 0.5, 1\}. 

\section{Experiment Details}
For all the benchmark datasets, we perform a 70/10/20 train/validation/test split to obtain the respective sets, consistent with previous research \cite{liu2023itransformer,haoyietal-informer-2021}. During the training process, we use the Adam optimizer with a learning rate of 1e-3 and employ the L1 loss function. The batch size is set to 128, and the default number of training iterations is set to 2,000. The sampling steps of ARMD are chosen from \{1, 2, 3, 4, 6, 8, 12\} via grid search, utilizing the validation set. Other diffusion-based TSF models use the default sampling steps of 100. The final evaluation is conducted on the test set. All experiments are conducted on a single NVIDIA GeForce RTX 4090 GPU with 24GB of memory.

\section{Data Description} \label{app:describe}
The experimental data encompasses seven widely used benchmark datasets, each distinguished by its unique set of characteristics:

Solar Energy \cite{lai2018modeling}: This dataset includes the solar power production records in the year of 2006, which is sampled every 10 minutes from 137 PV plants in Alabama State. The dataset contains 7,177 timesteps.

Exchange \cite{lai2018modeling}: This dataset includes daily exchange rates of eight countries from 1990 to 2016, with 8 variables and 7,588 timesteps per variable, displaying significant volatility. 

ETT Datasets \cite{haoyietal-informer-2021}: The ETT datasets consist of hourly-level data (ETTh1 and ETTh2) and 15-minute-level data (ETTm1 and ETTm2), with oil and load features of electricity transformers recorded from July 2016 to July 2018. Each dataset has seven variables, with ETTh datasets containing 17,420 timesteps and ETTm datasets containing 69,680 timesteps.

Stock \cite{yoon2019time}: This dataset contains the daily historical Google stocks data from 2004 to 2019, including as features the volume and high, low, opening, closing, and adjusted closing prices. The dataset contains 3,685 timesteps.

\begin{figure*}[!]
\centering
\includegraphics[width=1\linewidth]{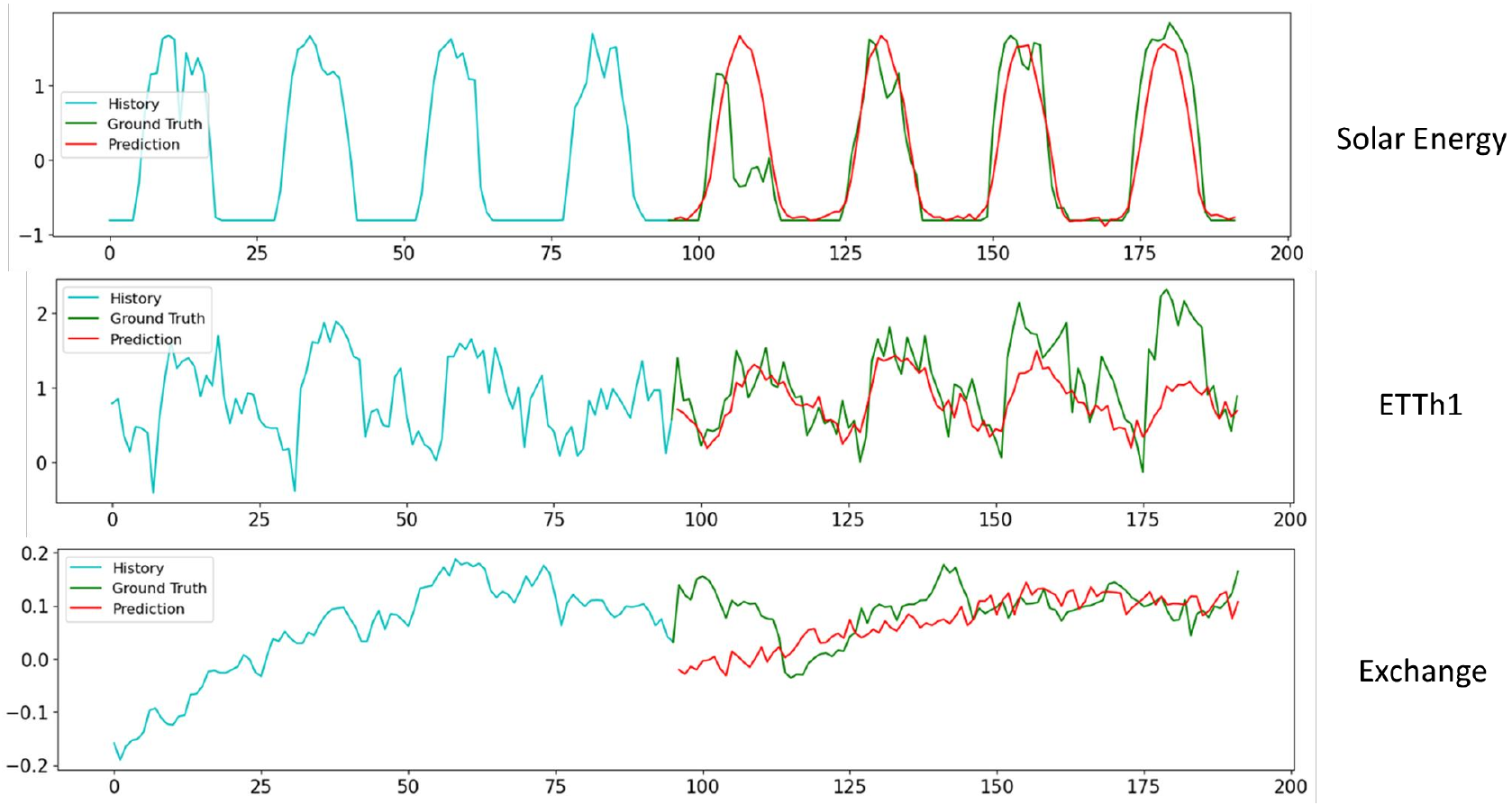} 
\caption{Prediction showcases of ARMD.}
\label{fig:showcases}
\end{figure*}

\section{Details on Comparison Models} \label{app:detail}

The descriptions and implementations of comparison models are detailed as follows:

Diffusion-TS \cite{yuan2024diffusion}: Diffusion-TS is a DDPM-based framework for generating high-quality multivariate time series samples. It utilizes an encoder-decoder Transformer with disentangled temporal representations and a decomposition technique to capture the semantic meaning of time series, while Transformers extract detailed sequential information from noisy inputs. The model excels in conditional generation tasks, such as forecasting and imputation. Their code can be found at {https://github.com/Y-debug-sys/Diffusion-TS}. 

MG-TSD \cite{fan2024mg}: MG-TSD addresses the instability of diffusion probabilistic models in TSF by leveraging the natural granularity levels within data. The model aligns the diffusion process with smoothing fine-grained data into coarse-grained representations, introducing a novel multi-granularity guidance diffusion loss function. This approach does not require external data and is versatile across various domains. The code is available at: {https://github.com/Hundredl/MG-TSD}.

TSDiff \cite{kollovieh2024predict}: TSDiff is a time series diffusion model that leverages a self-guidance mechanism for conditioning during inference, without the need for auxiliary networks or changes to the training procedure. This model excels across various time series tasks, including forecasting, refinement, and synthetic data generation. Their publicly available code is at {https://github.com/amazon-science/unconditional-time-series-diffusion}.

D3VAE \cite{li2022generative}: D3VAE is a bidirectional variational auto-encoder equipped with diffusion, denoising, and disentanglement, and it addresses TSF challenges by leveraging a coupled diffusion probabilistic model to augment data, integrating multiscale denoising score matching for accuracy, and enhancing interpretability and stability through disentangling multivariate latent variables. Their code is at: {https://github.com/PaddlePaddle/PaddleSpatial}.

TimeGrad \cite{timegrad}: TimeGrad is an auto-regressive model for multivariate probabilistic TSF that leverages diffusion probabilistic models to sample from the data distribution at each time step by estimating its gradient. By optimizing a variational bound on the data likelihood, TimeGrad learns these gradients and uses Langevin sampling during inference to convert white Gaussian noise into samples of the desired distribution. Their publicly available code is at {https://github.com/zalandoresearch/pytorch-ts}.



iTransformer \cite{litransformer}: iTransformer addresses the limitations of traditional Transformer models, which struggle with large lookback windows and fail to learn variate-centric representations. Instead of modifying the Transformer architecture, iTransformer applies attention and feed-forward networks on inverted dimensions. Time points are embedded into variate tokens to capture multivariate correlations, while each token is processed to learn nonlinear representations. This approach enhances performance and efficiency in TSF. Their code is available at this repository: {https://github.com/thuml/iTransformer}.

PatchTST \cite{nie2022time}: Adopting a segmental perspective on time series, PatchTST divides series into discrete time segments, each functioning as an independent token. PatchTST leverages attention mechanism to capture the dependencies between these tokens. The available code can be found at {https://github.com/yuqinie98/patchtst}.

DLinear \cite{Zeng2022AreTE}: DLinear employs a linear model to outperform most of Transformer-based models in TSF. Their publicly available source code can be found at {https://github.com/cure-lab/LTSF-Linear}.

TimesNet \cite{wu2023timesnet}: TimesNet introduces a novel paradigm by transforming 1D time series into a set of 2D tensors based on multiple periods. This transformation is followed by feature extraction utilizing a CNN. The code is available at \url{https://github.com/thuml/Time-Series-Library}.

Client \cite{gao2023client}: Client incorporates linear modules to learn trend information and employs an cross-variable Transformer module to capture cross-variable dependencies. In addition, it simplifies the embedding and position encoding layers within the cross-variable Transformer module and replaces the decoder module with a projection layer. The code for Client is available at \url{https://github.com/daxin007/Client}.

\section{Prediction Showcases} \label{app:showcase}
We present the prediction showcases of ARMD on different datasets, as shown in Fig. \ref{fig:showcases}. As can be seen from the figure, ARMD demonstrates excellent forecasting performance on time series with pronounced periodicity or trend characteristics, effectively capturing the underlying patterns in these time series.

\end{document}